\pdfoutput=1

\documentclass[11pt]{article}

\usepackage[preprint]{acl}

\usepackage{times}
\usepackage{latexsym}

\usepackage[T1]{fontenc}

\usepackage[utf8]{inputenc}

\usepackage{microtype}

\usepackage{inconsolata}

\usepackage{graphicx}
\usepackage{url}

\usepackage{setspace}
\usepackage{threeparttable}
\usepackage{supertabular}
\usepackage{bm}
\usepackage{amsthm}
\usepackage{mathrsfs}
\usepackage{siunitx}
\usepackage{footmisc}
\usepackage{footnote}
\usepackage{enumitem}
\usepackage{array}
\usepackage{CJK}
\usepackage{enumitem}
\usepackage{colortbl}
\usepackage{tikz}
\usepackage[edges]{forest}

\usepackage{lipsum}
\usepackage{booktabs}
\usepackage{longtable}
\usepackage{multirow}
\usepackage{algorithm}
\usepackage{algorithmic}
\usepackage{subcaption}
\usepackage{xcolor}
\usepackage{graphicx}
\usepackage{latexsym}
\usepackage{amsmath}
\usepackage{makecell}
\usepackage{fdsymbol}
\usepackage{pifont}
\usepackage{array}
\usepackage{fancyvrb}
\usepackage{tcolorbox}
\usepackage{tabularx}

\usetikzlibrary{trees,positioning,shapes,shadows,arrows.meta}

\usepackage[edges]{forest}
\definecolor{hidden-draw}{RGB}{0,0,0}
\definecolor{hidden-blue}{RGB}{194,232,247}
\definecolor{hidden-orange}{RGB}{243,202,120}
\definecolor{hidden-yellow}{RGB}{242,244,193}
\definecolor{tree-level-1}{RGB}{245,20,85}
\definecolor{tree-level-2}{RGB}{246,86,118}
\definecolor{tree-level-3}{RGB}{248,177,193}
\definecolor{tree-leaf}{RGB}{176,230,198}

\title{A Survey on Multi-Turn Interaction Capabilities of Large Language Models}

\author{Chen Zhang, Xinyi Dai, Yaxiong Wu,
Qu Yang \\ \textbf{Yasheng Wang}, \textbf{Ruiming Tang}, \textbf{Yong Liu} \\
  Huawei Noah’s Ark Lab \\
  \texttt{zhang.chen4@huawei.com}
}

\begin{document}
\maketitle
\begin{abstract}

Multi-turn interaction in the dialogue system research refers to a system's ability to maintain context across multiple dialogue turns, enabling it to generate coherent and contextually relevant responses. Recent advancements in large language models (LLMs) have significantly expanded the scope of multi-turn interaction, moving beyond chatbots to enable more dynamic agentic interactions with users or environments. In this paper, we provide a focused review of the multi-turn capabilities of LLMs, which are critical for a wide range of downstream applications, including conversational search and recommendation, consultation services, and interactive tutoring. This survey explores four key aspects: (1) the core model capabilities that contribute to effective multi-turn interaction, (2) how multi-turn interaction is evaluated in current practice,  (3) the general algorithms used to enhance multi-turn interaction, and (4) potential future directions for research in this field.

\end{abstract}

\section{Introduction}

The rapid growth in the field of large language models (LLMs)~\citep{openai-2024-gpt4,google-2024-gemini} has brought unprecedented opportunities to various downstream applications, including customer service~\citep{pandya-2023-automating}, healthcare~\citep{zhou-etal-2024-medicinellmsurvey}, virtual assistants~\citep{dong-etal-2023-intelligentassistant}, finance~\citep{li-etal-2023-llminfinance}, and more. Within these applications, LLMs are often deployed as interactive agents through chat interfaces, providing seamless communication and real-time responses to users' complex and multi-faceted queries. Such queries frequently require multiple rounds of interaction to fully address, underscoring the critical role of LLMs in managing ongoing conversations effectively~\citep{kwan-etal-2024-mt}. 

During the course of interaction with LLM-based agents, users expect them to consistently demonstrate capabilities, such as following instructions~\citep{chung-etal-2024-scaling}, memorizing and recalling information~\citep{zhang-etal-2024-surveyllmmemory}, reasoning through complex tasks~\citep{huang-etal-2024-planningsurvey}, and more. In this context, multi-turn interaction acts as a crucial bridge where the LLM’s core capabilities are seamlessly integrated, ultimately enhancing user satisfaction. While numerous surveys examine LLMs in a broad context~\citep{zhao-etal-2023-LLMSurvey, minaee-etal-2024-llmsurvey} or focus on specific capabilities such as instruction following~\citep{zhang-etal-2024-instructionsurvey}, reasoning~\citep{plaat-etal-2024-reasoning}, memory management~\citep{zhang-etal-2024-surveyllmmemory}, and planning~\citep{huang-etal-2024-planningsurvey}, there is currently no comprehensive survey dedicated to multi-turn interactions in LLMs.


Moreover, existing surveys on multi-turn dialogue systems~\citep{zhang-etal-2021-surveycomprehension, ni-etal-2023-recent, yi-etal-2024-survey} primarily focus on dialogue-specific tasks rather than examining LLMs as dynamic agents interacting with users and their environments. As multi-turn interactions powered by LLMs become increasingly crucial for handling more complex user tasks, our survey aims to provide a comprehensive overview of the current research in this area. We explore LLMs as dynamic agents that engage with both users and their environments to accomplish diverse tasks and improve the overall user experience. In particular, we review studies on model-specific capabilities that support and improve such interactions, offering valuable insights for both academic researchers and industry practitioners.

\paragraph{Scope}
While there is an extensive body of research on the use of LLMs in multi-turn dialogues across various applications, this survey does not cover all downstream applications related to multi-turn interactions, such as emotional support chatbots, medical consultations, and others. Instead, we focus on the model-specific capabilities essential for multi-turn interactions with both users and the environment. We place particular emphasis on works related to evaluation and algorithms for improving these model-specific capabilities. Finally, we primarily review papers from the LLM era, focusing on those came out since the end of 2022, when ChatGPT marked the beginning of chat-based LLMs, up to the present.

The subsequent content of our survey is structured into four sections. The first section reviews how LLM-based multi-turn interaction is evaluated (\S\ref{sec:evaluation}). The second section covers works on the essential LLM capabilities required for successful multi-turn interactions (\S\ref{sec:task-multi-turn}). The third section explores algorithms for enhancing multi-turn interaction (\S\ref{sec:method}). The final section concludes the survey and offers insights into future directions (\S\ref{sec:conclusion}).



\section{Multi-Turn Interaction Evaluation}
\label{sec:evaluation}

Evaluation is a driving force behind the advancement of LLM research. Through the use of benchmarks, evaluation tools, and metrics, it provides valuable insights, helping researchers and practitioners assess the relative performance of their proposals against state-of-the-art models. It also helps accelerate model iteration and identifies key areas for improvement. In this section, we review current practices for evaluating LLM-based multi-turn interactions, focusing on the various model-specific capabilities involved.

\subsection{General Multi-Turn Evaluation Framework}

The evaluation of User-LLM interactions focuses on assessing how well LLMs engage with users in multi-turn dialogues, emphasizing aspects such as naturalness, task completion, and user satisfaction. Existing research often encapsulates these elements under a unified concept of ``human preference'', which reflects the overall quality of the interaction from the user's perspective.

For instance,~\citet{zheng2023judging} introduce MT-Bench, a multi-turn question set, and Chatbot Arena, a crowd-sourced evaluation platform, to assess LLMs' alignment with human preferences in both controlled environment and real use-case scenarios, respectively. MT-Bench uses a rating-based approach to evaluate the LLMs' multi-turn generations, while Chatbot Arena employs a pairwise comparison method to assess the relative performance of two LLMs based on their interactions with real users. To reduce the costs of human evaluation and enhance reproducibility, the ``LLM-as-a-Judge'' framework is introduced, which leverages strong LLMs like GPT-4 to perform the rating and pairwise comparisons. They report high correlations between the LLM-generated judgments and human assessments. Since then, ``LLM-as-a-Judge'' has become a standard approach for evaluating user-LLM interactions. All the works discussed subsequently adopt the ``LLM-as-a-Judge'' to evaluate LLM performance.

MT-Bench++~\citep{sun-etal-2024-parrot} extends MT-Bench by adding six follow-up questions to each test instance, creating an eight-turn evaluation with a total of 80 sessions and 640 utterances. All the follow-up questions are manually crafted to be clear, fluent, and include ellipsis and anaphora, making them challenging for evaluating the multi-turn context understanding capability of LLMs.

Finally, MT-Bench-101~\citep{bai-etal-2024-mt} is by far the most comprehensive evaluation benchmark for multi-turn interaction. It employs a three-tier hierarchical taxonomy. The first tier includes perceptivity (understanding context), interactivity (proactively engaging users), and adaptability (responding effectively to user feedback). The second tier breaks these down into core capabilities: perceptivity consists of context memory, understanding, and interference; interactivity involves questioning; and adaptability includes reasoning, reflection, and paraphrasing. These capabilities are further divided into 13 specific tasks. Task-specific instruction templates prompt GPT-4 to gather multi-turn instructions, which are then used to assess LLMs' detailed multi-turn interaction abilities.






\subsection{Evaluating Instruction-Following}

\begin{table*}[!ht]	
\centering
\resizebox{\linewidth}{!}{
    \begin{tabular}{l|l}
    \toprule
    \textbf{Patterns}  & \textbf{Description} \\ \midrule
   Instruction Clarification & \makecell[l]{Users may provide initial instructions that are vague or ambiguous. Across \\ the course of interaction, the LLMs need to ask clarifying questions or request \\ additional information to fully understand the user's intent.} \\ \midrule
   Instruction Expansion & \makecell[l]{As the conversation progresses, users may provide supplementary details or \\ broaden the scope of the initial instruction.} \\ \midrule
   Constraint Addition & \makecell[l]{Users introduce new constraints or conditions in follow-up turns that the LLM \\ need to integrate into the ongoing task.} \\ \midrule
   Instruction Refinement & \makecell[l]{Over the course of interaction, the user may refine their initial request,  the LLM \\ need to be flexible and update its understanding as the instruction evolves.} \\ \midrule
    Global Instruction Consistency & \makecell[l]{The user requests LLMs to maintain a consistent tone, role, or style of communication \\ throughout the course of interaction.} \\
    \bottomrule
    \end{tabular}
}
\caption{User interaction patterns that necessitates multi-turn instruction-following.}
\label{tab:instruction-patterns}
\end{table*}

A core aspect of human preference is the ability to follow user instructions~\citep{zhou-etal-2023-ifeval}, particularly the consistent adherence to user's requests throughout the course of interaction. Evaluating multi-turn interactions requires designing tasks that assess a model's capability to understand and adhere to specific user-LLM interaction patterns. Table~\ref{tab:instruction-patterns} outlines five common user-LLM interaction patterns. For example, Multi-IF~\citep{he-etal-2024-multiif} primarily focuses on constraint addition. By combining multiple single-turn instructions from the IFEval benchmark~\citep{zhou-etal-2023-ifeval} into multi-turn sequential prompts, LLMs are expected to produce outputs that adhere to all the requirements specified by the users up to the current turn.
The tasks in Multi-IF focus specifically on verifiable instructions, such as "write a summary in 300 words" or "use all uppercase letters."

Moreover,~\citet{zhang-etal-2024-probing} investigates instruction clarification, assessing LLMs' ability to handle ambiguous queries. They adopt the 20-Questions entity-deduction game (Q20) to evaluate whether LLMs can accurately identify the correct entity. In this game, the LLMs ask the judge 20 different questions and make their guess based on the feedback provided.

Instead of targeting a single pattern, MT-Eval~\citep{kwan-etal-2024-mt} provide a more comprehensive assessment by decomposing multi-turn instruction following into fine-grained categories and addressing multiple interaction patterns. Ispects pertain to instruction expansion, refinement, and global instruction consistency in Table~\ref{tab:instruction-patterns}. 

\subsection{Evaluating Conversational Qualities}

Besides multi-turn instruction following, there are works on assessing the general conversational qualities of the multi-turn dialogues, such as coherence and human-likeness. For example,~\citet{duan-etal-2024-botchat} introduces the BotChat framework, in which different LLMs are iteratively prompted to generate multi-turn dialogues that resemble human-human conversations. A GPT-4 evaluator is then used to assess the number of turns generated by the LLMs within the dialogue. LLMs with lesser number of detected turns are deemed as exhibiting more human-likeness.

While BotChat only targets human-likeness, there are works on multi-dimensional evaluation. For example,~\citet{fu-etal-2024-gptscore} developed an automatic evaluator using GPT-3.5 to score different quality aspects of multi-turn dialogues based on the likelihood of the positive answers to predefined instruction prompts. They explore various evaluation settings, including zero-shot, in-context learning, and adding task-specific and dimension-specific instructions.~\citet{zhang-etal-2024-comprehensiveanalysis} provide a comprehensive analysis of over 30 LLMs in evaluating multi-turn dialogues, examining aspects such as coherence, engagement, and informativeness through prompting. They validate the findings in~\citet{zheng2023judging} that strong LLMs, such as GPT-4, produce evaluation results that are highly correlated with human evaluation.

\subsection{Evaluating Multi-Turn Reasoning} 

Beyond instruction-following and general conversational abilities, the capacity to reason across multiple rounds of interaction and recover from previous errors based on external feedback~\citep{kamoi-etal-2024-llms} is also a crucial aspect of multi-turn interaction capabilities in LLMs. In this context, we focus on multi-turn math reasoning, code reasoning, and general reasoning in LLMs.  

MathChat-Bench~\citep{liang-etal-2024-mathchat} is specifically designed to benchmark the multi-turn math reasoning abilities of various LLMs. It introduces four types of tasks: follow-up QA, error correction, error analysis, and problem generation. The test data for each task is generated by prompting GPT-4 according to task-specific designs. LLM performance on follow-up QA and error correction is evaluated by comparing the generated answers to the ground truth. For error analysis and problem generation, performance is assessed using the LLM-as-a-Judge approach~\citep{zheng2023judging}. 


\citet{yang2023intercode} highlight the significance of interacting with execution environments for LLM code reasoning and propose InterCode, a lightweight RL environment, where code execution is treated as actions and feedback as observations. InterCode integrates three interactive coding environments, Bash, SQL, and Python, using data from the static NL2Bash, Spider, and MBPP datasets respectively, to assess the code generation performance of different LLMs. The authors find that multi-turn prompting and interactive feedback enable the models to perform error correction and context discovery. This significantly enhances their performance compared to the single-turn baseline.

\citet{banatt2024wilt} highlight that LLMs struggle with multi-turn reasoning tasks requiring evidence gathering and logical conclusions, which should be analyzed separately from memory. They introduce the Wason Inductive Logic Test (WILT), a challenging benchmark designed to resist memorization, inspired by the Wason 2-4-6 task~\citep{wason1960onthefailure}. In WILT, LLMs identify an underlying rule by posing test cases and receiving feedback from the judge through up to 30 multi-turn interactions. The authors find that LLMs perform poorly on this task, with the best model achieving only 28\% accuracy, revealing a significant gap in their ability to handle complex reasoning. It is worth noting that scaling inference time compute, as demonstrated by OpenAI's O1, is likely to significantly enhance performance on single-turn multi-step reasoning tasks. However, its effectiveness for multi-turn reasoning tasks remains an area for further investigation in future research.

\subsection{Evaluating Multi-Turn Agentic Skills}

\citet{weng2023agent} defines LLM-powered autonomous agents as comprising an LLM core, planning capabilities, memory systems, and tool utilization. These skills are also crucial for executing successful multi-turn interactions with both the users or the environment. 

\paragraph{Multi-Turn Planning}
To evaluate multi-turn planning, \citet{xiao-etal-2024-flowbench} introduce FlowBench, a benchmark that integrates workflow-related knowledge in text, code, and flowchart formats, and generates realistic user-agent interactions using GPT-4 for both roles. When acting as users, GPT-4 relies on predefined profiles that include user backgrounds, goals, and contextual elements. These interactions serve as ground-truth test data for evaluating the planning capabilities of other LLMs. The authors use a simulated session framework, where GPT-4 interacts with the target LLM agent based on a task-user summary derived from the ground-truth session, which includes user goals and tool usage details. The performance of the LLM agent is measured by its task success rate and the number of sub-goals it successfully completes during the session. 


\paragraph{Multi-Turn Tool Use}
Over the course of multi-turn interaction, the integration of tools helps maintain context, streamline task execution, and provide more precise responses. Several works specifically target evaluating the multi-turn tool use of LLMs. For instance,~\citet{li-etal-2023-api} propose API-Bank, a benchmark designed to evaluates multi-turn tool use capabilities of LLMs by leveraging 73 commonly used APIs and 314 tool-use dialogues, which include 753 annotated API calls. API-Bank assesses the model's abilities in planning, retrieving, and calling API tools to fulfill user requirements. To reduce the high costs of manual annotation, the authors introduce Multi-agent, which utilizes LLMs to automatically generate tool-augmented training data at scale. Multi-agent consists of five collaborative agents that sequentially generate domains, APIs, user queries, API calls, and responses. Similarly, \citet{farn-etal-2023-tooltalk} propose ToolTalk, a benchmark consisting of 78 multi-turn user-agent dialogues across various tools in domains such as accounting, email, and calendar management. The dialogues are divided into easy and challenging sets to assess the LLMs' success rate in task execution.


\paragraph{Context Memory} 

During multi-turn interactions, LLM-based agents are expected to recall previous context and effectively integrate relevant information to address the current task.~\citet{maharana-etal-2024-evaluating} introduces LOCOMO, a benchmark for evaluating very long-term, open-domain, multimodal dialogues involving LLM agents. It aims to assess the ability of LLMs to handle dialogues with up to 600 turns and spanning 32 sessions. By integrating event graphs and personas, the study evaluates LLMs' memory, temporal reasoning, and multimodal interaction. The findings reveal that while long-context LLMs and RAG models show some improvement, they still significantly lag behind human performance in temporal reasoning and event graph summarization tasks. Additionally, LongMemEval~\citep{wu-etal-2024-longmemeval} is a benchmark designed to evaluate long-term memory in chat assistants. It tests five core abilities: information extraction, multi-session reasoning, knowledge updates, temporal reasoning, and abstention. With 500 questions based on dynamic multi-session user-assistant dialogues, it challenges models by requiring them to recall and reason across extended histories.


\tikzstyle{my-box}= [
    rectangle,
    draw=hidden-draw,
    rounded corners,
    text opacity=1,
    minimum height=1.5em,
    minimum width=5em,
    inner sep=2pt,
    align=center,
    fill opacity=.5,
]
\tikzstyle{leaf}=[my-box, minimum height=1.5em,
    fill=blue!15, text=black, align=left,font=\small,
    inner xsep=2pt,
    inner ysep
=4pt,
]
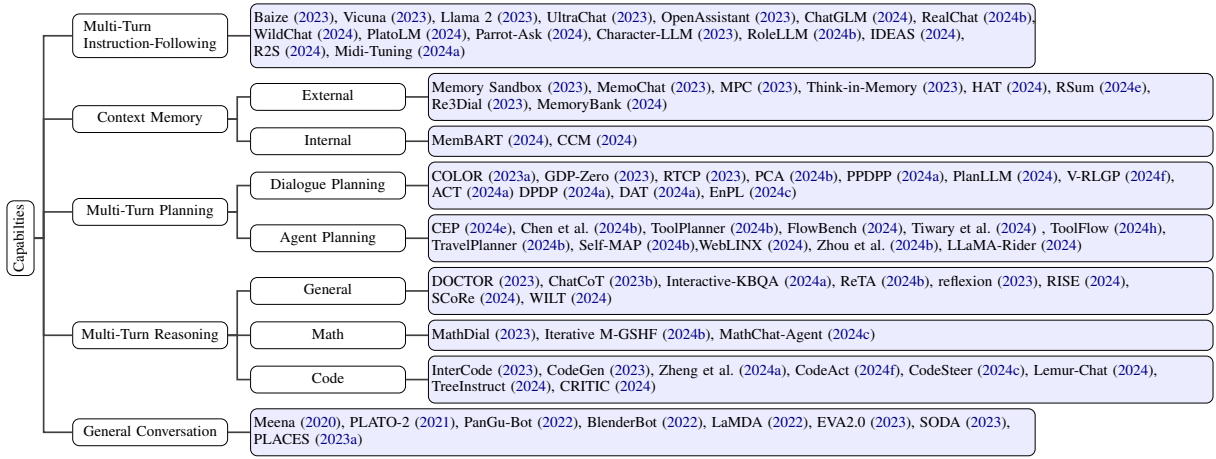
\begin{figure*}[ht]

    \centering
    \resizebox{\textwidth}{!}{
        \begin{forest}
            forked edges,
            for tree={
                grow=east,
                reversed=true,
                anchor=base west,
                parent anchor=east,
                child anchor=west,
                base=left,
                font=\small,
                rectangle,
                draw=hidden-draw,
                rounded corners,
                align=left,
                minimum width=4em,
                edge+={darkgray, line width=1pt},
                s sep=3pt,
                inner xsep=2pt,
                inner ysep=3pt,
                ver/.style={rotate=90, child anchor=north, parent anchor=south, anchor=center},
            },
            where level=1{text width=8.0em,font=\small,}{},
            where level=2{text width=8.0em,font=\small,}{},
            where level=3{text width=8.0em,font=\small,}{},
            where level=4{text width=8.0em,font=\small,}{},
[Capabilties, ver
    [Multi-Turn \\ Instruction-Following, text centered
        [Baize~\shortcite{xu-etal-2023-baize}{,} Vicuna~\shortcite{vicuna2023}{,} 
        Llama 2~\shortcite{touvron-etal-2023-llama2}{,} UltraChat~\shortcite{ding-etal-2023-enhancing}{,} OpenAssistant~\shortcite{kopf-etal-2023-openassistant}{,}
        ChatGLM~\shortcite{glm-etal-2024-chatglm}{,} RealChat~\shortcite{zheng2024lmsyschatm}{,} \\
        WildChat~\shortcite{zhao2024wildchat}{,}
            PlatoLM~\shortcite{kong-etal-2024-platolm}{,}
            Parrot-Ask~\shortcite{sun-etal-2024-parrot}{,}
            Character-LLM~\shortcite{shao-etal-2023-character}{,}
            RoleLLM~\shortcite{wang-etal-2024-rolellm}{,}
            IDEAS~\shortcite{ou-etal-2024-inductive}{,} \\
            R2S~\shortcite{hou-etal-2024-rawtext}{,}
            Midi-Tuning~\shortcite{wang-etal-2024-instruct}, leaf, text width=42em]
        ]
        [Context Memory, text centered
            [External, text centered
                [Memory Sandbox~\shortcite{huang-etal-2023-adjunct}{,}
                MemoChat~\shortcite{lu-etal-2023-memochat}{,}
                MPC~\shortcite{lee-etal-2023-prompted}{,}
                Think-in-Memory~\shortcite{liu-etal-2024-tim}{,}
                HAT~\shortcite{aadharsh-etal-2024-enhancing}{,}
                RSum~\shortcite{wang-etal-2024-recursively}{,} \\
                Re3Dial~\shortcite{wen-etal-2023-re3dial}{,}
                MemoryBank~\shortcite{zhong-etal-2024-memorybank},
                leaf, text width=42em]
            ]
            [Internal, text centered
                [MemBART~\shortcite{wu-yu-2024-stateful}{,} CCM~\shortcite{kim2024compressed},
                leaf, text width=42em]
            ]
        ]
        [Multi-Turn Planning, text centered
            [Dialogue Planning, text centered
                [COLOR~\shortcite{wang-etal-2023-dialogue}{,} GDP-Zero~\shortcite{yu-etal-2023-prompt}{,} RTCP~\shortcite{dao-etal-2023-reinforced}{,} PCA~\shortcite{li-etal-2024-pcaplanning}{,} PPDPP~\shortcite{deng2024plugandplay}{,}  PlanLLM~\shortcite{gloria-silva-etal-2024-plan}{,} 
                V-RLGP~\shortcite{zhang-etal-2024-probing}{,} \\ 
                ACT~\shortcite{chen-etal-2024-act}
                DPDP~\shortcite{he-etal-2024-planning}{,}
                DAT~\shortcite{li-etal-2024-dat}{,} EnPL~\shortcite{zheng-etal-2024-thoughts}, leaf, text width=42em]
            ]
            [Agent Planning, text centered
                [CEP~\shortcite{zhang-etal-2024-ask}{,}  
                Chen et al.~\shortcite{chen-etal-2024-relyllmagentsdraft}{,} ToolPlanner~\shortcite{wu-etal-2024-toolplanner}{,}
                FlowBench~\shortcite{xiao-etal-2024-flowbench}{,}
                Tiwary et al.~\shortcite{tiwary-etal-2024-from} {,} 
                ToolFlow~\shortcite{wang-etal-2024-toolflow}{,}
                \\
                TravelPlanner~\shortcite{chen-etal-2024-relyllmagentsdraft}{,}
                Self-MAP~\shortcite{deng-etal-2024-multi}{,}WebLINX~\shortcite{lu2024weblinx}{,}
                 Zhou et al.~\shortcite{zhou-etal-2024-enhancing}{,} LLaMA-Rider~\shortcite{feng-etal-2024-llama},
                leaf, text width=42em]
                ]
            ]
        [Multi-Turn Reasoning, text centered
            [General, text centered
                [DOCTOR~\shortcite{chae-etal-2023-dialogue}{,}
                ChatCoT~\shortcite{chen-etal-2023-chatcot}{,}
                Interactive-KBQA~\shortcite{xiong-etal-2024-interactive}{,} ReTA~\shortcite{duan-etal-2024-reta}{,} 
                reflexion~\shortcite{shinn2023reflexion}{,}
                RISE~\shortcite{qu2024recursive}{,} \\
                SCoRe~\shortcite{kumar-etal-2024-score}{,}
                WILT~\shortcite{banatt2024wilt}, leaf, text width=42em]                
            ]
            [Math, text centered
                [MathDial~\shortcite{macina-etal-2023-mathdial}{,} Iterative M-GSHF~\shortcite{xiong2024building}{,} 
                MathChat-Agent~\shortcite{wu-etal-2024-mathchat},
                leaf, text width=42em]
            ]
            [Code, text centered
                [InterCode~\shortcite{yang2023intercode}{,} CodeGen~\shortcite{nijkamp2023codegen}{,} Zheng et al.~\shortcite{zheng-etal-2024-whatmakesllm}{,} CodeAct~\shortcite{wang-etal-2024-executablecode}{,} CodeSteer~\shortcite{chen-etal-2024-steeringllms}{,}  Lemur-Chat~\shortcite{xu2024lemur}{,} \\
                TreeInstruct~\shortcite{kargupta-etal-2024-instruct}{,}
                CRITIC~\shortcite{gou2024critic},
                    leaf, text width=42em]
                ]
            ]
        [General Conversation, text centered
            [Meena~\shortcite{adiwardana-etal-2020-humanlike}{,} 
            PLATO-2~\shortcite{bao-etal-2021-plato}{,}
            PanGu-Bot~\shortcite{mi-etal-2022-pangubot}{,}
            BlenderBot~\shortcite{shuster-etal-2022-blenderbot3}{,}
            LaMDA~\shortcite{thoppilan-etal-2022-lamda}{,} 
            EVA2.0~\shortcite{Gu-etal-2023-eva2}{,} 
            SODA~\shortcite{kim-etal-2023-soda}{,}
            \\
            PLACES~\shortcite{chen-etal-2023-places}, leaf, text width=42em]
        ]
    ]
\end{forest}
}
    \caption{Relevant works on multi-turn interaction organized by capabilities.}
    \label{fig:organize-work-by-capability}
\end{figure*}



\paragraph{Combination of Agentic Skills}

Effective multi-turn interactions often depend on the seamless integration of multiple skills, rather than the reliance on a single isolated ability. Evaluations of such interactions are typically conducted by treating the LLM as an agent and assessing its ability to successfully execute tasks by synergistically utilizing these skills. For example,~\citet{wang2024mint} highlight the importance of leveraging interactions among users, LLMs, and external tools when solving complex tasks. Noting the absence of a suitable evaluation benchmark for such interactions, the authors introduce MINT, a framework designed to assess LLMs' ability to solve tasks through multi-turn interactions with environment. In MINT, the LLM can use external tools by generating and executing Python code or refine its solutions by collecting natural language feedback from the GPT-4 user simulator. Data from eight existing datasets, covering tasks like reasoning, code generation, and decision-making, are restructured into a user-environment interaction format, yielding 586 representative multi-turn instances. 


Additionally,~\citet{liu2024agentbench} introduces AgentBench, a multi-dimensional benchmark consisting of eight distinct environments across code, web, and gaming domains. It is designed to evaluate the reasoning, planning, and instruction-following capabilities of LLMs as agents. AgentBench unifies a variety of agent-based tasks, including interacting with operating systems and web shopping, and features practical, multi-round interaction challenges. For each task, the estimated number of rounds required to solve a problem ranges from 5 to 50.

To assess agent performance across a diverse set of tasks and in partially observable settings, where agents must actively explore and understand their environment, \citet{ma2024agentboard} propose the AgentBoard benchmark. This benchmark includes 9 unique tasks and 1,013 typical environments, spanning embodied AI, game agents, networked agents, and tool-based agents. Each task-specific data sample involves multi-round interactions, with sub-goals annotated for the task. A progress rate metric is used to track the agent's detailed progress in achieving these sub-goals, offering a more granular evaluation than existing benchmarks, which typically rely solely on final task success rates.

\section{Core LLM Capabilities in a Multi-Turn Context}
\label{sec:task-multi-turn}

After reviewing how multi-turn interactions of LLMs are evaluated in the existing literature, we highlight works that enhance key model-specific capabilities for more effective multi-turn interactions. While these capabilities have been extensively studied in existing research, there is a lack of comprehensive overview on how they function and differ in multi-turn interaction scenarios, which contrast with the more prevalent focus on single-turn instruction tasks in current literature. In this section, we highlight five high-level capabilities relevant to multi-turn interaction and Figure~\ref{fig:organize-work-by-capability} provides an overview of the relevant works organized by the capabilities.



%


\subsection{Multi-Turn Instruction Following}
\label{subsec:instruction-following}

Existing works on LLM instruction tuning~\citep{wang-etal-2023-self-instruct,chung-etal-2024-scaling} primarily work on QA-style instruction data, without considering prior interactions or building context over multiple turns. However, when dealing with more complex tasks, users do not necessarily provide a one-time instruction, but exhibit different interaction patterns that necessitates multi-turn instruction following~\citep{bai-etal-2024-mt,kwan-etal-2024-mt}.


Multi-turn instruction following is often not directly targeted during the instruction-tuning process of many open-source LLMs~\citep{xu-etal-2023-baize, vicuna2023, touvron-etal-2023-llama2, glm-etal-2024-chatglm}. Instead, this capability is acquired implicitly through supervised fine-tuning on large-scale instruction datasets~\citep{ding-etal-2023-enhancing, kopf-etal-2023-openassistant}, which usually contain a mix of single-turn and multi-turn conversational data. However, datasets that explicitly capture the interaction patterns outlined in Table~\ref{tab:instruction-patterns} remain relatively scarce.


There are also works curating multi-turn interaction data in the wild to study how real users interact with the LLMs. For example,~\citet{zheng2024lmsyschatm} introduce LMSYS-Chat-1M dataset, which contain one million conversations between real users and 25 LLMs. The dataset is collected via their hosted Vicuna demo and Chatbot Arena websites.~\citet{zhao2024wildchat} studies how real-world users prompt ChatGPT and make available a WildChat corpus comprising 1 million user-ChatGPT conversations, comprising over 2.5 million interaction turns. To our knowledge, no existing work has systematically analyzed and extracted interaction data specifically designed for multi-turn instruction following from publicly available resources. While \citet{shachar-etal-2024-learning} and \citet{shi-etal-2024-wildfeedback} attempt to mine feedback signals from real user-LLM interactions to generate preference data for aligning language models optimized for single-turn instruction following, future research could explore methods for extracting high-quality data relevant to multi-turn instruction following.

In addition to gathering real user-LLM multi-turn interactions,~\citet{kong-etal-2024-platolm} and~\citet{sun-etal-2024-parrot} propose employing LLM-based user simulators to emulate human users interacting with ChatGPT. This approach is used to generate the multi-turn datasets, SocraticChat and Parrot-40K, respectively. During their data collection,~\citet{sun-etal-2024-parrot} explicitly target the "ambiguous instruction" interaction pattern by training their parrot-ask user simulator to ask questions containing anaphora or ellipsis while the user simulator in~\citet{kong-etal-2024-platolm} adopts a socratic questioning approach to guide the multi-turn generations of ChatGPT. 

However,~\citet{ou-etal-2024-inductive} critiques that LLM-based user simulators often capture common dialogue flow patterns from the training data, leading them to generate more generic instructions. To address this issue, the authors propose a two-phase induction-deduction framework, IDEAS. In the induction phase, GPT-4 is prompted to extract high-level instruction strategies from real user-LLM instructional interactions. During the deduction stage, the user simulator selects an appropriate instruction strategy based on the new dialogue context and generates an instruction conditioned on that strategy. A system agent then responds to the instruction. This process repeats iteratively to build a multi-turn instructional dialogue. It is worth noting that the instruction strategies, derived from the vast amount of real user-LLM multi-turn data, capture a diverse set of user interaction patterns. These strategies hold significant potential for improving multi-turn instruction following and present an exciting research direction for further exploration. 

Character-LLM~\citep{shao-etal-2023-character} and RoleLLM~\citep{wang-etal-2024-rolellm} are examples of "role consistency" within the "global instruction consistency" category of Table~\ref{tab:instruction-patterns}. They require LLMs to respond in a manner consistent with the behavior of specific characters and to maintain this consistency throughout the interaction. In contrast, \citet{touvron-etal-2023-llama2} advocates for a broader form of multi-turn consistency, which ensures adherence to instructions specified in the system message. Furthermore,~\citet{wang-etal-2024-instruct} introduces multi-round interactive dialogue tuning (Midi-Tuning) framework to enhance role consistency of the LLMs. It utilizes two adapters on top of LLMs to model the agent and the user separately. The adapters are fine-tuned using their respective role-specific utterances in dialogues, alternating round by round, through an utterance-level memory caching mechanism.


\subsection{Context Memory}
\label{subsec:context-memory}

Context memory is fundamental for maintaining continuity in multi-turn interactions, allowing relevant information from previous exchanges to be retained and utilized throughout the conversation. We categorize works on enhancing context memory into external and internal memory mechanisms.


\subsubsection{External Memory}

One line of research studies augmenting LLMs with different memory management mechanisms to help them track and recall information during the course of interaction. For instance,~\citet{huang-etal-2023-adjunct} design an interactive memory sandbox to store the dialogue history as memory objects. Users can view and manipulate these objects, performing operations such as addition, deletion, modification, and summarization. ChatGPT memory\footnote{\url{https://openai.com/index/memory-and-new-controls-for-chatgpt/}} is a concrete implementation of the "memory sandbox".

\citet{aadharsh-etal-2024-enhancing} propose to store dialogue history using a "Hierarchical Aggregate Tree" (HAT) memory structure, where salient information is stored in the tree nodes. The text in each parent node is aggregated from the texts of its child nodes, with ChatGPT assisting in the information aggregation process. When responding to a current query, ChatGPT acts as a tree traversal agent, navigating the HAT by generating an optimal sequence of traversal actions based on the text representation at each node and the user query. The traversal continues until ChatGPT determines that sufficient information has been gathered to answer the query. 

RSum~\citep{wang-etal-2024-recursively} employs a sequential process for tracking dialogue history. Initially, an LLM is prompted to generate a summary based on a short context. The model then updates the summary by incorporating a short span of subsequent context. This process is repeated iteratively, with each new context refining the summary, until it captures all the key information from the entire dialogue history.

\citet{liu-etal-2024-tim} proposes Think-in-Memory (TiM), a method designed to mimic how humans remember and selectively recall thoughts. After each interaction turn, the LLM reasons over its response and saves key points, referred to as "thoughts", into a hash-based memory storage. In subsequent turns, the model retrieves relevant thoughts from memory to respond to user queries. The hash-based storage not only enables fast memory retrieval but also supports memory updates, maintaining a detailed index of historical thoughts.

The scarcity of long-session dialogues also challenges effective long-range context retention. To address this, \citet{wen-etal-2023-re3dial} propose $\text{Re}^3\text{Dial}$, a three-stage approach (Retrieve, Reorganize, Rescale) that generates large-scale long-session dialogue data. The method retrieves coherent, consecutive dialogue sessions using a contrastive learning-based retriever to capture semantic and discourse relations. A diversity sampling strategy ensures minimal repetition, while iterative concatenation of short-turn dialogues builds extended conversations, supporting the external memory capability required for maintaining context in long-session interactions.

In summary, external memory mechanisms significantly enhance LLMs' ability to maintain continuity in multi-turn interactions. Techniques like interactive memory sandboxes, hierarchical aggregate trees, recursive summarization, and hash-based memory storage enable efficient tracking, storage, and retrieval of dialogue history, thereby ensuring contextually appropriate and consistent responses and enriching user interaction experiences.

\subsubsection{Internal Memory}

Another line of research focuses on enhancing "internal memory," where contextual information is stored directly as part of the language model's internal modules. For example, building upon Memformer~\citep{wu-etal-2022-memformer},~\citet{wu-yu-2024-stateful} introduces MemBART, a stateful, memory-augmented transformer encoder-decoder model. MemBART maintains dialogue history as internal memory hidden states and interacts with this memory through a memory reader and writer, which are modules integrated into the modified transformer architecture. To update memory states at each layer, a separate transformer model is employed. Additionally, a residual gated update mechanism is introduced to control the extent to which existing memory is retained or overwritten at each time step.

\citet{kim2024compressed} proposes CCM, a compressed context memory system. CCM dynamically updates during inference by optimizing a lightweight conditional LoRA\citep{hu2022lora}, enabling the model to create a compressed attention key-value memory of contextual information during the forward pass. To mitigate training inefficiencies from the recursive context compression for dynamic updates, the authors propose unrolling the procedure and processing it in parallel using concatenation or merging functions. During the inference, the model conditioned on the compressed memory when responding to the user's current query. 

There are fewer works along the second research line compared to the first, likely due to the complexity of modifying the model architecture while maintaining effective memory management and preserving the model's generation capabilities. Furthermore, as the language models scale up, the plug-and-play nature of the first research line enables memory mechanisms to be more efficient and easier to modify. 

\subsection{Planning}
\label{subsec:planning}

Following our discussion on context memory, we now address multi-turn planning, an essential skill for LLM-based agents. This involves organizing, prioritizing, and adapting responses over time to maintain coherence, relevance, and goal-directed dialogue. As tools are integrated, planning complexity increases, necessitating effective management of additional layers of information and decision making. We review existing works on multi-turn planning based on the dialogue planning and agent planning categories.

\subsubsection{Dialogue Planning}

The dialogue planning subsection includes works on shaping the overall structure and flow of the conversation between the agent and the user. A key challenge in this area is effectively managing the trajectory of conversations over time,~\citet{wang-etal-2023-dialogue} addresses it by a global planning mechanism based on a Brownian bridge stochastic process~\citep{revuz1999continuous}, which maps dialogue topics or action-topic pairs into a latent space. This process is conditioned on the current dialogue context and the specified goal.

To address the challenges of global dialogue planning, especially the challenge of simulating future interactions and training on large annotated datasets for look-ahead algorithms like Monte Carlo Tree Search (MCTS)~\citep{yang-etal-2021-multi},~\citet{yu-etal-2023-prompt} propose GDP-Zero, which formulates dialogue planning as a stochastic Markov Decision Process (MDP) and utilizes Open-Loop MCTS at decision time. Open-Loop MCTS mitigates compounding errors by continuously re-generating responses during the tree search. Few-shot prompting with ChatGPT is used for various operations, such as simulating user and system responses, evaluating task progress, and predicting the next dialogue act, eliminating the need for training data.

While global planning addresses long-term conversation structure, proactive dialogue planning focuses on dynamically adjusting the conversation to keep users engaged and ensure smooth interaction. A notable approach in this area is RTCP~\citep{dao-etal-2023-reinforced}, which integrates long-term and short-term planning through a balanced gating mechanism. Long-term planning, formulated as a MDP and trained using an actor-critic framework, encourages the system to focus on achieving predefined goals, such as recommending targeted items, while avoiding overly prolonged conversations. In contrast, short-term planning uses a knowledge-integrated multi-head attention mechanism to predict the next action tuple (a topic and sub-goal pair) based on the dialogue context, associated knowledge, and previous actions, ensuring user engagement and conversational smoothness.

\citet{deng2024plugandplay} proposes PPDPP, a tunable plug-and-play policy planner that determines actions at each conversation step. These actions are translated into natural language instructions to guide the LLM in generating the next response. A vanilla policy gradient RL algorithm optimizes the planner using self-play dialogues from two LLMs and scoring from a third LLM.

To overcome the efficiency issues of prompt engineering and the suboptimal performance of existing policy networks,~\citet{he-etal-2024-planning} applies the dual-process theory of human cognition~\citep{kahneman2003maps} to dialogue planning, distinguishing between System 1 (intuitive, fast responses) and System 2 (analytical, slow reasoning). A neural policy LM (System 1) generates quick responses to familiar scenarios, while a MCTS-based policy planner (System 2) handles complex, novel situations. A two-stage training is applied whereby the model is first trained using offline RL on static dialogue data, then guided by MCTS self-play simulations. The learning process helps balance efficiency and strategic depth.

Moreover, to improve plan rationality and reduce the dependency on human simulations in tree-search policy learning,~\citet{zheng-etal-2024-thoughts} propose to distill natural language plans from existing target-driven conversation corpus and then transfer to new dialogue scenarios with demonstration-guided in-context learning. 


To address ambiguous user intents by asking clarification questions,~\citet{chen-etal-2024-act} introduce the Action-Based Contrastive Self-Training (ACT) framework. ACT is designed to enhance LLMs' ability to select appropriate conversational strategies in multi-turn and ambiguous contexts. The framework aligns with the goal of modeling actions within conversations by employing a sample-efficient, quasi-online Direct Preference Optimization (DPO) algorithm that prioritizes winning responses corresponding to the correct action over those derived from negatively sampled dialogue actions. For example, when facing ambiguous user request, the appropriate action is to seek clarification rather than to answer.

Overall, dialogue planning integrates various innovative methods to steer conversation trajectories, including global planning mechanisms, proactive strategies, plug-and-play policy planners, and dual-process theories. Recent developments also focus on enhancing plan rationality and addressing ambiguous user intents with frameworks like ACT.

\subsubsection{Agent Planning}

While dialogue planning specifically addresses the conversational aspects of interaction, agent planning has a broader scope that includes all agent behaviors and actions. These actions include tasks such as tool use, information retrieval, and other forms of interaction. This section reviews key works on agent planning in the context of multi-turn interactions. Most studies focus on the interactions between LLM-based agents and the environment, particularly in tasks involving multi-turn tool use and environment feedbacks.

In real-world tool-use scenarios, users often describe their needs without explicitly specifying the tools required, making it necessary for LLMs to break down complex instructions into subtasks. These agents determine and interact with the appropriate tools over multiple rounds based on each subtask and ultimately provide a reasonable answer. This process aligns with the task decomposition planning skill, as defined by \citet{huang-etal-2024-planningsurvey}.

To acquire such a skill,~\citet{wu-etal-2024-toolplanner} introduce ToolPlanner, a two-stage reinforcement learning framework. The first stage of supervised fine-tuning in ToolPlanner involves tag extraction, high-level solution paths generation, and ToolPlanner interacting with external tools to construct a solution tree, where each interaction, comprising thought generation, API requests, and observations, forms an intermediate node. The final answer is derived as the rightmost path of the solution tree. The RL stage employs the RRHF approach~\citep{yuan2023rrhf}, using a combination of task completion and instruction-following rewards to score candidate solutions, with the higher-scoring solution used for continual fine-tuning of ToolPlanner.

Furthermore,~\citet{wang-etal-2024-toolflow} propose ToolFlow, a data synthesis pipeline for creating multi-turn tool-use dialogues to support supervised fine-tuning of LLMs. ToolFlow employs a graph-based sampling strategy to generate relevant tool combinations and a planned-generation strategy to create dialogue plans, ensuring the synthesis of coherent dialogues that effectively integrate these tool combinations.

A concrete multi-turn tool use scenario is the conversational web navigation task, which requires LLM-based agents to plan and interact with websites to address user requests. For example,~\citet{lu2024weblinx} proposes the WebLINX benchmark, which includes 100K interactions across 150+ real-world websites, to investigate whether LLM-based agents can replicate human behavior when navigating the web. To address the limited context length of LLMs and the context-dependency challenges in multi-turn web navigation,~\citet{deng-etal-2024-multi} introduce Self-Reflective Memory-Augmented Planning (Self-MAP), which leverages memory utilization and self-reflection techniques.

In addition to tool use, planning skills are also essential when LLMs interact with open-world environments.~\citet{feng-etal-2024-llama} investigate how LLM-based agents can continuously acquire environmental knowledge and adapt in such settings. Using Minecraft, an open-ended sandbox world, as their environment, the authors propose LLaMA-Rider, a multi-round feedback-revision mechanism that guides LLMs to select appropriate revision actions based on real-time environmental feedback. Additionally, they incorporate sub-task relabeling to help LLMs maintain consistency in their planning and better understand the combinatorial relationships between tasks.

Another significant development in agent planning is Proactive Agent Planning, as introduced by~\citet{zhang-etal-2024-ask}, a task where LLM-powered agents predict clarification needs based on user-agent conversations and agent-environment interactions. These agents must invoke external tools to gather information and generate a plan to meet the user's requirements. The authors propose a multi-agent framework, Clarification-Execution-Planning (CEP), which features three specialized agents for the clarification, execution, and planning tasks respectively. 

In summary, agent planning significantly broadens the scope of LLM-based interactions by integrating complex behaviors such as tool use and environmental adaptability. Techniques like ToolPlanner and ToolFlow illustrate the potential for structured, multi-turn tool interactions, while approaches such as Self-MAP and LLaMA-Rider emphasize the importance of memory and real-time feedback in dynamic environments. Proactive planning, as seen in the CEP framework, underscores the necessity for agents to anticipate and clarify user needs dynamically. These advancements not only enhance the agents' operational efficiency but also pave the way for more intuitive and context-aware interactions, ultimately pushing the boundaries of what LLM-based agents can achieve in real-world applications.

\subsection{Multi-Turn Reasoning}
\label{subsec:reasoning}

Reasoning-intensive tasks, such as mathematics and coding, require LLMs to analyze question context, draw inferences, and apply logical frameworks to generate accurate solutions~\citep{huang-chang-2023-towards}. 
Recent advances in LLM prompting techniques, such as Chain-of-Thought~\citep{wei-etal-2022-cot} and Tree of Thoughts~\citep{yao2023tree}, encourage LLMs to follow human problem-solving patterns, significantly enhancing their reasoning abilities. However, for more complex tasks, intermediate feedback from the users or external tools may be necessary to refine reasoning and ensure accuracy~\citep{lightman2023letsverifystepstep}. 
In such scenarios, multi-turn interaction can help effectively incorporate such feedback in the reasoning process of LLMs, by allowing them to maintain reasoning flow and adapt based on new information or clarifications from external sources of validation. 

In this section, we categorize relevant works on LLM multi-turn reasoning into three groups: general reasoning, math, and code, followed by a review of representative studies in each category.

\subsubsection{General Reasoning}

ChatCoT~\shortcite{chen-etal-2023-chatcot} models the chain-of-thought (CoT) reasoning process as multi-turn conversations, integrating tools in the natural course of interaction. The process begins by providing the LLMs with essential background knowledge, including descriptions of available tools, relevant task examples, and demonstrations of the decomposed chain-of-thought. Next, the LLM iteratively uses the tools to assist with reasoning at each turn. The process continues until all reasoning steps are completed and the final answer is derived.

\citet{xiong-etal-2024-interactive} proposes the interactive-KBQA framework to handle complex user queries. In this framework, the LLM acts as an agent, interacting with the knowledge base to decompose the problem-solving process into a series of multi-turn LLM-KB interactions. At each turn, the LLM reasons and generates an appropriate action to interact with the knowledge base (KB) using a set of specific tools. These tools return execution results, which serve as observations for further reasoning. The process continues iteratively until the final answer is reached. The authors also introduce a human-model collaborative mechanism, allowing humans to refine the LLM's intermediate reasoning and actions at the end of each turn.

\paragraph{Strategic Reasoning}

To examine the strategic multi-turn reasoning capabilities of LLMs,~\citet{duan-etal-2024-reta} analyze their behavior in complete- and incomplete-information games, such as Tic-Tac-Toe and Texas Hold'em Poker. An online racing setting is adopted to compare various LLM-based reasoning agents, including random, CoT, CoT with Self-Consistency (CoT-SC)~\citep{wang2023selfconsistency}, Tree of Thoughts (ToT)~\citep{yao2023tree}, and ReAct~\citep{yao2023react}. It is found that advanced reasoning agents show only slight improvements over the random agent, which randomly select the next action during its interaction with the environment. The authors perform error analysis using offline probing data, which includes targeted questions with verified ground truth to identify common reasoning errors, such as hallucination. Based on these insights, they propose Recursively Thinking Ahead (ReTA), to enhance LLM performance in strategic multi-turn reasoning tasks.

\paragraph{Commonsense Reasoning}

Commonsense reasoning is another crucial component of general reasoning skills. In multi-turn interactions, one significant challenge is multi-hop commonsense reasoning, where key evidence required for reasoning is dispersed across and implicitly expressed in multiple dialogue turns. To address this issue, \citet{chae-etal-2023-dialogue} propose Dialogue CoT, which frames multi-hop commonsense reasoning as a CoT reasoning process~\citep{wei-etal-2022-cot}. In Dialogue CoT, LLMs are prompted to iteratively generate questions and answers related to the dialogue context, using commonsense relations from the ATOMIC knowledge base~\citep{hwang-etal-2021-atomic}. These answers are combined as the corresponding rationale for guiding response generation. Through this information-seeking process, the LLMs identify relevant contextual cues and infer underlying knowledge necessary for generating appropriate responses.

\paragraph{Self Correction}

An important trait of LLMs' general reasoning skill is the ability to self-correct. Prompting techniques can be designed to facilitate self-correction behavior in LLMs across multiple turns. A representative work is reflexion~\citep{shinn2023reflexion}, an approach that uses verbal reinforcement to enhance agents' learning by reflecting on prior mistakes. Reflexion converts feedback from the environment (binary or scalar) into textual summaries, which are then provided as additional context in subsequent turns to guide the agent’s improvement.

Several studies argue that prompting alone is insufficient to enable self-correction abilities in large language models (LLMs)~\citep{welleck2023generating, qu2024recursive, zhang-etal-2024-small}. These works advocate for supervised fine-tuning as a more effective approach to instill such capabilities in LLMs. In the context of multi-turn self-correction, \citet{qu2024recursive} propose the RISE approach, which models the response refinement process as a multi-turn Markov Decision Process (MDP) using a single-turn dataset of prompts and oracle responses. The policy LLM generates an initial response, then a fixed user feedback is used to prompt the LLM to reflect and revise its response. The procedure repeats until the oracle response is reached. To ensure continuous improvement of the policy responses along the interaction Multitrajectory, the authors propose two mechanisms. First, using a stronger teacher LLM to produce improved responses conditioned on the previous interaction trajectory. Another mechanism is to sample the Best-of-N response from the policy model at each state. The collected data is used to fine-tune the policy using a reward-weighted
regression objective~\citep{jan-etal-2007-rwr}.





\citet{kumar-etal-2024-score} claim that purely supervised fine-tuning (SFT) on offline model-generated multi-turn correction data is insufficient for instilling self-correction behavior in models. This insufficiency arises due to two main issues: (1) a distribution mismatch between the mistakes made by the data-collection policy and the model's own responses, and (2) behavior collapse, where the model learns to favor a limited correction strategy that is hard to generalize. Hence, they propose SCoRe, an on-policy multi-turn RL framework extending ArCHer~\citep{zhou-etal-2024-archer}. SCoRe employs a two-stage learning approach to promote self-correction behavior. In Stage 1, the model is initialized by decoupling its behavior across two attempts\footnote{Attempt refers to the response to answer the user's query.} in the multi-turn dialogue, optimizing the accuracy of the latter attempt while constraining the distribution of first attempt to match that of the base model. Stage 2 then jointly optimizes the reward for both attempts, with a bias in the reward function to prevent collapse to the direct solution and instead reinforce progress in self-correction.

In summary, effective multi-turn general reasoning combines foundational general reasoning methods, strategic planning, commonsense knowledge, and self-correction. It ensures logical consistency, allows for long-term problem-solving, and grounds interactions in real-world context.

\subsubsection{Math Reasoning}

LLMs have shown strong performance in solving mathematical problems, especially in single-turn question-answering tasks. However, in real-world applications like interactive chatbots and problem-solving assistants, math tasks go beyond simple single-turn questions and demand more sophisticated abilities, such as understanding multi-turn interaction, reasoning through problems step by step, and providing educational feedback. In this context, we review relevant works that address these challenges. For instance,~\citet{wu-etal-2024-mathchat} introduce the MathChat-Agent framework, featuring an LLM agent tasked with problem-solving and a user proxy agent responsible for executing tools and providing feedback. The two agents engage in multi-turn interactions to collaboratively solve math problems. Different multi-step tool-use and reasoning techniques are integrated into problem-solving turns of interaction process, such as ReAct~\citep{yao2023react}, ART~\citep{paranjape-etal-2023-art}, and ToolFormer~\citep{schick2023toolformer}.

Additionally,~\citet{macina-etal-2023-mathdial} introduce the MathDial dataset to advance research on automatic dialogue tutors. MathDial consists of 3,000 one-on-one teacher-student tutoring dialogues, each grounded in multi-step math reasoning problems. The dataset is created by pairing human teachers with an LLM, which is prompted to simulate common student errors in solving math reasoning problems. The authors also observe that LLMs like GPT-3 excel at problem-solving, they often struggle with tutoring, as they may provide factually incorrect feedback or prematurely reveal solutions to students. To address this issue, the authors guide teachers to provide learning opportunities by steering students through a series of scaffolding questions, grounded in a structured taxonomy of teacher moves. The data collected from this approach can be used to fine-tune LLMs, improving their ability to function as effective tutors.

While most existing work focuses on generating synthetic multi-turn math reasoning data and fine-tuning LLMs using these datasets,~\citet{xiong2024building} propose the Iterative M-GSHF approach, which complements existing methods and further enhances the multi-turn math reasoning capabilities of LLMs via direct preference learning. The learning process is framed as a Markov decision process (MDP), taking into account interactions with the external environment. Two variants of multi-turn direct alignment based on preference algorithms, M-DPO and M-KTO, are developed by adapting existing DPO~\citep{rafailov2023direct} and KTO~\citep{ethayarajh-etal-2024-kto} algorithms to incorporate trajectory preferences. Additionally, the online iterative alignment mechanism~\citep{yuan-etal-2024-selfrewarding, zhang-etal-2024-ts} is integrated with M-DPO and M-KTO to further improve alignment performance. Although the work mainly targets multi-turn math reasoning capability, it can easily generalize to other multi-turn interaction tasks.

\subsubsection{Code Reasoning}

Multi-turn interactions plays a pivotal role in tasks involving programming assistance, debugging, and code generation. These interactions are not limited to a single-step solution but evolve through a series of exchanges, where the model iteratively refines or adjusts code based on user feedback, problem clarification, or new requirements. This section explores recent representative works on how multi-turn interactions are applied to code-related tasks.

CodeGen~\citep{nijkamp2023codegen} highlight a multi-turn program synthesis approach, in which users engage with the model over multiple interactions. In each turn, users provide natural language specifications, prompting the LLM to generate synthesized subprograms. Through this iterative process, the human and LLM collaboratively construct the entire program. Using this approach, the authors create a multi-turn program synthesis benchmark to explore how scaling laws influence multi-turn programming abilities in LLMs. They evaluate models of varying sizes, trained sequentially on THEPILE, BIGQUERY, and BIGPYTHON datasets, finding that performance improves with larger models and more data. Multi-turn synthesis consistently outperforms single-turn instructions across all model scales.

Rather than relying on human collaboration,~\citet{kargupta-etal-2024-instruct} presents TreeInstruct, a method for Socratic-style code debugging using LLMs. Unlike existing teacher LLMs that offer direct solutions, TreeInstruct engages students in multi-turn interactions, asking probing questions that help them identify and resolve code errors. By utilizing a dynamic state-space planning algorithm, it tailors its questions to the student’s current knowledge state and responses.

Besides including execution feedback into multi-turn code generation,~\citet{zheng-etal-2024-whatmakesllm} introduce CoT reasoning prompts, such as those involving predicting problem attributes or generating natural language solutions, prior to providing execution feedback from intermediate results to LLMs. The authors also propose sampling CoT-enhanced multi-turn code generation trajectories from an LLM and applying rejection sampling fine-tuning to further improve the LLM's performance. Similarly, CodeSteer~\citep{chen-etal-2024-steeringllms} also explores the integration of textual reasoning and code execution feedback for multi-turn code refinement, demonstrating performance improvements across six models. However, both strategies are constrained by the LLM’s intrinsic textual reasoning capability and the frequency of code interpreter utilization, which may lead to increased runtime and longer output sequences.

\subsection{General Conversation}
\label{subsec:multi-turn-dialogue}

Multi-turn interaction has a long history in dialogue system research~\citep{cui-etal-2020-mutual,xu_zhao_zhang-2021-topicaware,zhang-etal-2021-dynaeval,ye-etal-2022-multiwoz,ni-etal-2023-recent}. Significant research efforts have focused on pre-training language models on human conversational data, leading to the development of human-like chatbots such as Meena~\shortcite{adiwardana-etal-2020-humanlike}, BlenderBot~\shortcite{shuster-etal-2022-blenderbot3}, LaMDA~\shortcite{thoppilan-etal-2022-lamda}, and others. With the scaling laws~\citep{kaplan2020scalinglaws} and instruction fine-tuning~\citep{chung-etal-2024-scaling}, the conversational capabilities of LLMs, such as coherence, naturalness, engagingness, and informativeness, have reached near-human proficiency, particularly in proprietary models~\citep{openai-2024-gpt4, geminiteam2024gemini}, which now effortlessly pass the Turing tests~\citep{jones2024people}.

Recent research focused on enhancing the multi-turn conversational capabilities of open-source LLMs primarily involves developing data pipelines to curate high-quality dialogue datasets. For example,~\citet{kim-etal-2023-soda} propose distilling social conversations from the closed-source GPT-3.5 model~\citep{ouyang2022training} by contextualizing social commonsense knowledge from the $\text{Atomic}^{10x}$ symbolic commonsense knowledge graph~\citep{west-etal-2022-symbolic}. The process involves retrieving social commonsense knowledge from the $\text{Atomic}^{10x}$, converting this knowledge into sentence form and generating a narrative based on the sentence, and inferring the conversation participants from the narrative to derive a conversation grounded in that narrative. In this way, the authors curate a million-scale, broad spectrum of social conversations, named SoDA. In a similar vein,~\citet{chen-etal-2023-places} propose using a small set of expert-written conversations as in-context exemplars to synthesize a social conversation dataset through prompting LLMs.


\section{Multi-Turn Interaction Algorithms}
\label{sec:method}

In this section, we discuss representative algorithms for enhancing multi-turn interactions without referencing any model-specific capabilities.

\citet{zhou-etal-2024-archer} propose the ArCHer framework to address the challenge of single-turn reinforcement learning (RL), where most approaches struggle to enable LLMs to perform credit assignment or reason about past actions across multiple turns. ArCHer employs a hierarchical RL strategy with two parallel algorithms: a high-level off-policy value-based RL to aggregate rewards across multiple utterances, and a low-level policy gradient RL that leverages the high-level value function to train token-level policies within each turn.

Additionally,~\citet{shani2024multiturn} extend the single-turn reinforcement learning from human feedback (RLHF) to multi-turn dialogues. They propose the Multi-turn Preference Optimization (MTPO) algorithm, which combines the Mirror Descent (MD) method~\citep{BECK2003167} with self-play techniques~\citep{silver-etal-2017-mastering}. MTPO is theoretically proven to converge to a Nash equilibrium~\citep{Nash1951} and utilizes a preference-based Q-function that accounts for the long-term effects of individual actions, incorporating preferences over entire conversations rather than just single-turn feedback. This approach outperforms RLHF baselines in tasks requiring long-term planning, such as education dialogue tasks.

To address the policy-induced covariate shift in multi-turn conversations without the need for an additional critic network, which is used in ArCHer and MTPO,~\citet{gao-etal-2024-refuel} introduce REFUEL to streamlines policy optimization by using the reparameterization trick~\citep{degrave-etal-2019-quinoa} to directly regress future returns (Q-values) based on the logarithm of policy ratios. REFUEL leverages the difference in conversation-level rewards from a shared prefix to estimate Q-value differences at the point where conversations diverge and it iteratively generates on-policy data composed of conversation prefixes and independent responses from the current policy, ensuring that the model trains on relevant conversational data it will encounter in real interactions. 

Due to the complexity of modeling the reward function and the instability of the RLHF training process,~\citet{shi-etal-2024-direct} extend the DPO algorithm~\citep{rafailov2023direct} to a multi-turn alignment setting. To solve the technical challenges of the inability to cancel the partition function in the original DPO formulation and the length disparities between preferred and dispreferred multi-turn trajectorie, the authors propose to incorporate a State-Action Occupancy Measure (SAOM) constraint into the RL objective and applying length normalization to the Bradley-Terry model, and then derive a loss function tailored for direct multi-turn preference alignment.

\section{Conclusion and Future Directions}
\label{sec:conclusion}

We highlight several directions worth exploring, categorized into three perspectives: training data, evaluation, and algorithms.

\paragraph{More Diverse Training Data}

Many of the works reviewed in this survey rely on prompting advanced LLMs to curate their multi-turn training data. However, this approach can result in less diverse and potentially biased datasets that do not accurately reflect the dynamics of real user-LLM interactions. Therefore, future research may focus on effectively mining and filtering high-quality data from real user-LLM interaction data, which provide authentic conversational patterns, diverse linguistic styles, and varied contextual nuances. Additionally, developing tailored user simulators to engage with LLMs, rather than merely prompting them to generate multi-turn data, could provide more realistic and varied interaction scenarios.

\paragraph{Calibrating LLM-based Evaluators} 

The LLM-as-a-Judge\footnote{The judge is mainly the most advanced GPT-4 model.}~\citep{zheng2023judging} paradigm is widely utilized to assess the performance of the LLMs due to its efficiency, reproducibility, simplicity, and strong correlation with human judgments. However, utilizing advanced LLMs for evaluation purposes is known to introduce various biases~\citep{wang-etal-2024-large-language-models-fair, dubois2024lengthcontrolled}. These biases can affect the fairness and accuracy of the evaluation outcomes, potentially favoring certain models or responses over others. Future research may improve the reliability of LLM-based evaluators by fine-tuning them with human-annotated data and implementing bias mitigation strategies. Additionally, integrating multiple evaluators and cross-validating with human judgments can enhance the robustness and validity of the evaluation process.

\paragraph{Generate Test Data Aligned with Real-User Satisfaction}

Existing approaches predominantly develop task-specific benchmarks to evaluate the multi-turn capabilities of LLMs. Test data are typically generated by powerful LLMs, such as GPT-4 or adapted from established benchmarks. These synthetic data may not sufficiently represent the diverse range of multi-turn interaction capabilities required for comprehensive assessment. Such data may lack the variability in conversational contexts, linguistic styles, and user intents that occur in real-world interactions. Future research may develop methods to create more diverse and representative test datasets by incorporating real user interaction logs, applying data augmentation techniques, and ensuring broad conversational scenarios. Researchers and practitioners may also consider designing task-specific interaction patterns, similar to those outlined in Table~\ref{tab:instruction-patterns} to guide the data collection process.

\paragraph{Self-Evaluation} 

Learning to reflect is a key capability in multi-turn interaction and self-evaluation is essential to learning to reflect, where the LLMs criticize their own responses to identify inconsistencies, gaps in information, or areas lacking clarity~\citep{yuan2024selfrewardinglanguagemodels}. By doing so, the model can adjust its subsequent responses to better align with the ongoing dialogue and user expectations.

\paragraph{Feedback Integration} 

Another intriguing research question is how to effectively incorporate user or environmental feedback to guide the generation of the next-turn response. In a multi-turn setting, user or environment feedback, whether explicit or implicit, can be incorporated into the model's reflective process. For instance, if a user expresses confusion or dissatisfaction with a particular response, the model can recognize these cues and adapt its future interactions to address the underlying issues~\citep{zhang2024aligninglanguagemodelsusing}. This dynamic adjustment enhances the model's ability to maintain user engagement and satisfaction over extended dialogues.




\paragraph{Multi-Turn Reasoning}

Lots of recent research efforts focus on single-turn multi-step reasoning~\citep{wei-etal-2022-cot,zhang2023automatic,lightman2024lets,wang-etal-2024-math}, leaving multi-turn reasoning under-explored, future research may explore implementing hierarchical dialogue management, which allows models to decompose complex conversations into manageable sub-tasks, facilitating systematic and structured reasoning. Modular and composable architectures can also be implemented with the integration of specialized reasoning modules for handling specific aspects of interaction, thereby improving the depth and accuracy of model responses.





\bigskip
\noindent
In conclusion, we have systematically reviewed the multi-turn interaction of large language models by examining evaluation practices, core capabilities, general algorithms, and future research directions. Our insights pave the way for developing more effective LLM-based multi-turn agents that better serve user needs and enhance overall satisfaction.




\bibliography{anthology,custom}




\end{document}